\documentclass[pdflatex,sn-mathphys-ay]{sn-jnl}

\usepackage{graphicx}%
\usepackage{multirow}%
\usepackage{amsmath,amssymb,amsfonts}%
\usepackage{amsthm}%
\usepackage{mathrsfs}%
\usepackage[title]{appendix}%
\usepackage{xcolor}%
\usepackage{textcomp}%
\usepackage{manyfoot}%
\usepackage{booktabs}%
\usepackage{algorithm}%
\usepackage{algorithmicx}%
\usepackage{algpseudocode}%
\usepackage{listings}%

\theoremstyle{thmstyleone}%
\newtheorem{theorem}{Theorem}

\newtheorem{proposition}[theorem]{Proposition}%

\theoremstyle{thmstyletwo}%
\newtheorem{remark}{Remark}%

\theoremstyle{thmstylethree}%

\theoremstyle{definition}

\newtheorem{assumption}{Assumption}

\begin{document}

\title[Article Title]{Measuring and Analyzing Intelligence via Contextual Uncertainty in Large Language Models using Information-Theoretic Metrics}

\author[1,2,3]{\fnm{Jae Wan} \sur{Shim}}\email{jae-wan.shim@kist.re.kr}

\affil[1]{\orgdiv{Extreme Materials Research Center}, \orgname{Korea Institute of Science and Technology}, \orgaddress{\street{5 Hwarang-ro 14-gil, Seongbuk}, \city{Seoul}, \postcode{02792},  \country{Republic of Korea}}}

\affil[2]{\orgdiv{Climate and Environmental Research Institute}, \orgname{Korea Institute of Science and Technology}, \orgaddress{\street{5 Hwarang-ro 14-gil, Seongbuk}, \city{Seoul}, \postcode{02792},  \country{Republic of Korea}}}

\affil[3]{\orgdiv{Division of AI-Robotics}, \orgname{KIST Campus, University of Science and Technology}, \orgaddress{\street{5 Hwarang-ro 14-gil, Seongbuk}, \city{Seoul}, \postcode{02792},  \country{Republic of Korea}}}

\abstract{Large Language Models (LLMs) excel on many task-specific benchmarks, yet the mechanisms that drive this success remain poorly understood. We move from asking what these systems can do to asking how they process information. Our contribution is a task-agnostic method that builds a quantitative Cognitive Profile for any model. The profile is built around the Entropy Decay Curve---a plot of a model's normalised predictive uncertainty as context length grows. Across several state-of-the-art LLMs and diverse texts, the curves expose distinctive, stable profiles that depend on both model scale and text complexity. We also propose the Information Gain Span (IGS) as a single index that summarises the desirability of a decay pattern. Together, these tools offer a principled way to analyse and compare the internal dynamics of modern AI systems.}

\keywords{Neural Networks, Large Language Models, Information Theory, Conditional Entropy, Uncertainty Quantification, Memorisation}

\maketitle

\section{Introduction}
Intelligence defies a single, agreed-upon definition \cite{neisser1979concept, sternberg2004culture}.  
Yet recent LLMs display behaviours that look intelligent, reaching human-level scores on complex tasks.  
If the concept itself is elusive \cite{bommasani2021opportunities, bender2021dangers}, how can we understand systems that seem to embody it?

LLMs give us a unique opening.  
For any context, they output a full probability distribution over the next token \cite{brown2020language, bishop2006pattern}.  
These distributions, derived from the model's internal logits, let us study information processing directly.

We ground our analysis in Shannon's information theory \cite{shannon1948}.  
Prior work has used information measures to trace information flow in deep nets \cite{tishby2000information, shwartz2017opening} and to gauge LLM confidence \cite{kadavath2022language, jiang2020can}.  
We build on this by noting that intelligent behaviour balances two extremes: rigid determinism, which lacks creativity, and pure randomness, which is noise.  
A capable system should lock in on clear evidence yet stay flexible when context is thin.  
We call this capacity \textbf{Adaptive Predictive Modulation}.

To measure it, we compute two entropy-based quantities for a fixed context length $k$:
\begin{itemize}
    \item \textbf{Average conditional entropy} $h_k$: residual uncertainty after seeing $k$ tokens.
    \item \textbf{Entropy of the average distribution} $H_k$: diversity of possible outputs averaged over all contexts of length $k$.
\end{itemize}
(Definitions follow in Section~\ref{sec:definitions}).  
From these we form the \textbf{Length-Conditional Uncertainty Index}
\[
   u_k := \frac{h_k}{H_k},\qquad 0 \le u_k \le 1,
\]
which places the model on a determinism---randomness scale for each $k$.

By tracing $u_k$ against $k$, we obtain the \textbf{Entropy Decay Curve}---a quantitative Cognitive Profile\footnote{Perplexity is typically derived from cross-entropy on ground-truth next tokens, whereas 
$h_k$ here is the predictive entropy of the model's full next-token distribution. Our index $u_k$ additionally normalises by $H_k$, which PPL does not capture.}.  
The curve's starting point, slope, and floor capture a model's information-handling strategy.
We present an empirical study of these profiles for several leading LLMs, showing how they reveal both genuine capability and issues such as data contamination \cite{carlini2022quantifying, achiam2023gpt}.

The rest of the paper defines the metrics, describes the experimental design, and analyses the resulting profiles.

\section{Definitions and Metrics}
\label{sec:definitions}
We begin by formalising the probability distributions produced by a Large Language Model (LLM) and the information-theoretic measures derived from them.  
These definitions underlie all subsequent analyses of predictive uncertainty.

\subsection{Predictive Distribution of an LLM}

Consider an autoregressive LLM that, given a context of \(k\) tokens, outputs a probability distribution over its full vocabulary.  
Let
\[
   X = (t_1, t_2, \dots, t_k)
\]
be such a context and let \(\mathcal{Y}\) denote the vocabulary (typically \(|\mathcal{Y}| \approx 128\,256\)).  
The model returns the conditional distribution
\[
   p(Y \mid X),
\]
where, for any token \(y \in \mathcal{Y}\),
\[
   p(y \mid X) = \Pr\bigl(\text{next token}=y \mid X\bigr).
\]

Crucially, \(p(Y \mid X)\) refers to the entire probability vector, not just a sampled outcome.  
The vector is obtained by applying the softmax function to the model's logits, the standard way to convert raw scores into a valid probability distribution in multi-class generation tasks \cite{goodfellow2016deep, vaswani2017attention}.

\subsection{Empirical Estimation of Entropies}
We estimate conditional and marginal entropies for a fixed context length \(k\) by sampling
\(N\) context windows \(\{x_1, x_2, \dots, x_N\}\) from a representative corpus.

\subsubsection{Conditional Entropy \(\boldsymbol{(h_k)}\)}

For each context \(x_i\) we compute its conditional entropy
\begin{align*}
   h_{x_i} &:= H(Y\mid X=x_i) \\
           &= -\sum_{y\in\mathcal{Y}} p(y\mid x_i)\,
                   \log_2 p(y\mid x_i).
\end{align*}
The \emph{average conditional entropy} for length \(k\) is the sample mean
\[
   h_k := \frac{1}{N}\sum_{i=1}^{N} h_{x_i},
\]
which represents the model's mean predictive uncertainty after reading \(k\) tokens.

\subsubsection{Marginal Entropy \(\boldsymbol{(H_k)}\)}

Exact marginal entropy requires averaging over all contexts and is intractable \cite{cover1999elements}.  
Instead, we form the \emph{average predictive distribution}
\[
   \bar{p}_k(y) := \frac{1}{N}\sum_{i=1}^{N} p(y\mid x_i),
   \qquad y\in\mathcal{Y},
\]
and define the empirical marginal entropy
\[
   H_k := -\sum_{y\in\mathcal{Y}} \bar{p}_k(y)\,
                   \log_2 \bar{p}_k(y).
\]
By concavity of entropy,
\[
H_k = H(\bar p_k)\;\ge\;\frac{1}{N}\sum_{i=1}^N H\!\bigl(p(\cdot\mid x_i)\bigr)
\;=\;\frac{1}{N}\sum_{i=1}^N h_{x_i}
\;=\;h_k,
\]
so \(H_k\) serves as an upper bound for \(h_k\).

\subsection{The Cognitive Profile: Uncertainty Index and Entropy Decay Curve}

\paragraph{Length-Conditional Uncertainty Index}
For each \(k\) we form
\begin{equation}
   u_k := \frac{h_k}{H_k}, \qquad 0 \le u_k \le 1,
\end{equation}
called the \emph{Length-Conditional Uncertainty Index}.  
It rescales residual uncertainty by the model's potential output diversity at that context length.

\paragraph{Entropy Decay Curve (EDC)}
Plotting \(u_k\) against \(k\) (e.g., \(k\in\{3,9,30,\dots\}\)) yields the \emph{Entropy Decay Curve}.  
The curve's starting height, rate of decline, and eventual plateau together provide a quantitative \emph{Cognitive Profile} of the model's information-processing behaviour.

\section{Experimental Setup}
This section describes the models, corpora, and procedures used to estimate the Entropy Decay Curves (EDCs).

\subsection{Models}

To compare information processing across architectures and sizes, we test three public LLMs.  
All are quantised to GGUF format with \texttt{Q4\_K\_M} weights for consistent, memory-efficient inference:
\begin{itemize}
    \item \textbf{Llama 3.3 70.6B} --- 70.6-billion parameters;
    \item \textbf{DeepSeek-R1 8.19B} --- 8.19-billion parameters in the Qwen3 line \cite{qwen3};
    \item \textbf{Qwen 2.5 7.62B} --- 7.62-billion parameters in the Qwen2 family \cite{qwen2.5}.
\end{itemize}

\subsection{Corpora}

All evaluations use three public-domain texts from Project Gutenberg\footnote{\url{https://www.gutenberg.org/ebooks/11}} \footnote{\url{https://www.gutenberg.org/ebooks/4300}} \footnote{\url{https://www.gutenberg.org/ebooks/48433}}
:
\begin{enumerate}
    \item \emph{Alice's Adventures in Wonderland} --- analysis begins at ``CHAPTER I. Down the Rabbit-Hole...''.
    \item \emph{Ulysses} --- analysis begins at ``-- I -- [1] Stately, plump Buck Mulligan...''.
    \item \emph{Kant's Critique of Judgement} --- analysis begins at ``PREFACE We may call the faculty of cognition from principles...''.
\end{enumerate}
Project Gutenberg headers, licences, and tables of contents are removed; the main text is left unchanged.

\subsection{Implementation Details and Procedure}

\paragraph{Software stack.}
\begin{itemize}
    \item \texttt{llama-cpp-python} for quantised inference
    \item \texttt{NumPy} for array operations
    \item \texttt{scipy.special} for numerically stable \texttt{softmax} and entropy
\end{itemize}

\paragraph{Sliding-window evaluation.}
\begin{itemize}
    \item \textbf{Window lengths} \(k \in \{3, 9, 30, 90, 300, 600\}\)
    \item \textbf{Samples per \(k\)} \(N = 1000\)
    \item \textbf{Tokens needed} \(1600 = \max(k) + N\)
    \item \textbf{Model context} \texttt{n\_ctx} \(= 2048\) with full GPU off-load (\texttt{n\_gpu\_layers} \(= -1\))
\end{itemize}

\paragraph{Algorithm.}
For each model and each window size \(k\):
\begin{enumerate}
    \item \textbf{Tokenise} the first 1600 tokens of the chosen corpus.
    \item For \(i = 1 \dots N\):
          \begin{enumerate}
              \item Reset the model state.
              \item Extract context \(x_i = (t_i,\dots,t_{i+k-1})\).
              \item Run the model on \(x_i\) to obtain logits.
              \item Compute the conditional entropy \(h_{x_i}\); accumulate \(p(Y\mid X=x_i)\) into a running average to form \(\bar{p}_k\).

          \end{enumerate}
    \item After \(N\) iterations compute
          \begin{itemize}
              \item \(h_k\) --- mean conditional entropy,
              \item \(H_k\) --- entropy of the averaged distribution,
              \item \(u_k = h_k / H_k\) --- Length-Conditional Uncertainty Index.
          \end{itemize}
\end{enumerate}
The full pipeline is repeated for every model and every \(k\), producing the data required to plot each Cognitive Profile.

\subsection{Positional Robustness Check (Positional Homogeneity)}
\label{ssec:positional_robustness}

A potential concern in sliding-window entropy estimation is that averages at different configurations
may implicitly mix different target positions in the source text, and thus reflect positional
non-stationarity rather than a pure context-length effect. To empirically probe this issue, we ran
blockwise positional robustness checks on the \textit{Alice's Adventures in Wonderland} corpus using
\textbf{Llama 3.3} at two context lengths: a short window (\(k=3\)) and a long window (\(k=600\)).

We select three disjoint target blocks indexed by \(j_0\in\{600,1200,1800\}\). For each block and each
\(k\in\{3,600\}\) we form
\[
X^{(k)}_j=(t_{j-k},t_{j-k+1},\dots,t_{j-1}),\qquad Y_j=t_j,\qquad j\in\{j_0,\dots,j_0+999\},
\]
compute the conditional entropy \(H(Y_j\mid X^{(k)}_j)\) and the corresponding predictive
distribution \(p(\cdot\mid X^{(k)}_j)\) for \(N=1000\) targets, then aggregate to obtain
\(h_k\), \(H_k\), and \(u_k=h_k/H_k\).

\begin{table}[t]
\centering
\caption{Positional robustness on \textit{Alice's Adventures in Wonderland} for \textbf{Llama 3.3} at two fixed context lengths (\(k=3\) and \(k=600\)). Each row averages over \(N=1000\) consecutive targets in the indicated range. Reported values are the mean conditional entropy \(h_k\), the entropy of the average predictive distribution \(H_k\), and the uncertainty index \(u_k=h_k/H_k\).}
\label{tab:positional_robustness_k3_k600}
\begin{tabular}{r r r r r r r}
\toprule
\multirow{2}{*}{\textbf{Target range \(j\)}} &
\multicolumn{3}{c}{\textbf{\(k=3\)}} &
\multicolumn{3}{c}{\textbf{\(k=600\)}} \\
\cmidrule(lr){2-4}\cmidrule(lr){5-7}
& \(\boldsymbol{h_3}\) (bits) & \(\boldsymbol{H_3}\) (bits) & \(\boldsymbol{u_3}\) &
\(\boldsymbol{h_{600}}\) (bits) & \(\boldsymbol{H_{600}}\) (bits) & \(\boldsymbol{u_{600}}\) \\
\midrule
600--1599  & 11.5641 & 13.3657 & 0.8652 & 0.1161 & 7.7403 & 0.0150 \\
1200--2199 & 11.6408 & 13.4568 & 0.8650 & 0.1194 & 7.7721 & 0.0154 \\
1800--2799 & 11.7438 & 13.4985 & 0.8700 & 0.1323 & 7.7415 & 0.0171 \\
\midrule
\textbf{Range (max--min)} &
0.1797 & 0.1328 & 0.0050 &
0.0162 & 0.0318 & 0.0021 \\
\bottomrule
\end{tabular}
\end{table}

Across these three within-text blocks, the uncertainty index remains stable at both short and long
context lengths as in Table~\ref{tab:positional_robustness_k3_k600}. For \(k=3\), \(u_3\) varies only from \(0.8650\) to \(0.8700\) (absolute range \(0.0050\),
i.e.\ \(\approx 0.58\%\) relative to the minimum). For \(k=600\), \(u_{600}\) varies from \(0.0150\) to
\(0.0171\) (absolute range \(0.0021\)). Overall, these results suggest that within the tested portion
of the corpus, positional effects are small relative to the magnitude of the entropy-based quantities,
supporting the use of block-averaged estimates for the main experiments. These positional checks are
auxiliary control experiments and therefore use a longer contiguous segment of the corpus than the
1600-token segment used in the main EDC measurements.

\subsection{Empirical Data and Initial Observations}
Tables~\ref{tab:entropy_ratio_ulysses}, \ref{tab:entropy_ratio_alice}, and \ref{tab:entropy_ratio_kant} report $h_k$, $H_k$, and the derived uncertainty index $u_k$ for all models on \textit{Ulysses}, \textit{Alice's Adventures in Wonderland}, and \textit{Kant's Critique of Judgement}, respectively.  The primary metric $u_k$ expresses residual prediction uncertainty normalised by the model's potential output diversity at each context length $k$.

\begin{table*}[ht]
\centering
\caption{Conditional entropy $h_k$, average‐distribution entropy $H_k$ (bits), and their ratio $h_k/H_k$  for each model on the \textit{Ulysses} corpus.}
\label{tab:entropy_ratio_ulysses}
\resizebox{\textwidth}{!}{%
\begin{tabular}{l l r r r r r r}
\toprule
\textbf{Model} & \textbf{Metric} & \textbf{k=3} & \textbf{9} & \textbf{30} & \textbf{90} & \textbf{300} & \textbf{600} \\
\midrule
\multirow{3}{*}{Llama 3.3 70.6B}
  & $h_k$      & {\small 11.2112} & {\small 5.4613} & {\small 1.9958} & {\small 0.5029} & {\small 0.2583} & {\small 0.2160} \\
  & $H_k$      & {\small 13.5514} & {\small 10.5765} & {\small 9.1482} & {\small 8.3146} & {\small 8.1838} & {\small 8.1527} \\
  & \textbf{$h_k/H_k$} & \textbf{0.8273} & \textbf{0.5164} & \textbf{0.2182} & \textbf{0.0605} & \textbf{0.0316} & \textbf{0.0265} \\
\midrule
\multirow{3}{*}{DeepSeek-R1 8.19B}
  & $h_k$      & {\small 5.7411} & {\small 4.6142} & {\small 3.9565} & {\small 3.7438} & {\small 3.6196} & {\small 3.5586} \\
  & $H_k$      & {\small 10.2484} & {\small 9.9338} & {\small 9.6758} & {\small 9.4896} & {\small 9.3911} & {\small 9.3497} \\
  & \textbf{$h_k/H_k$} & \textbf{0.5602} & \textbf{0.4645} & \textbf{0.4089} & \textbf{0.3945} & \textbf{0.3854} & \textbf{0.3806} \\
\midrule
\multirow{3}{*}{Qwen2.5 7.62B}
  & $h_k$      & {\small 6.3178} & {\small 5.0146} & {\small 4.3945} & {\small 4.2200} & {\small 3.8902} & {\small 3.6960} \\
  & $H_k$      & {\small 10.7054} & {\small 10.0853} & {\small 9.7528} & {\small 9.5351} & {\small 9.3776} & {\small 9.2664} \\
  & \textbf{$h_k/H_k$} & \textbf{0.5902} & \textbf{0.4972} & \textbf{0.4506} & \textbf{0.4426} & \textbf{0.4148} & \textbf{0.3989} \\
\bottomrule
\end{tabular}}
\end{table*}

\begin{table*}[ht]
\centering
\caption{Conditional entropy $h_k$, average‐distribution entropy $H_k$ (bits), and their ratio $h_k/H_k$  for each model on the \textit{Alice's Adventures in Wonderland} corpus.}
\label{tab:entropy_ratio_alice}
\resizebox{\textwidth}{!}{%
\begin{tabular}{l l r r r r r r}
\toprule
\textbf{Model} & \textbf{Metric} & \textbf{k=3} & \textbf{9} & \textbf{30} & \textbf{90} & \textbf{300} & \textbf{600} \\
\midrule
\multirow{3}{*}{Llama 3.3 70.6B}
  & $h_k$      & {\small 11.6270} & {\small 5.0797} & {\small 1.4042} & {\small 0.1720} & {\small 0.1086} & {\small 0.1161} \\
  & $H_k$      & {\small 13.4119} & {\small 10.0430} & {\small 8.4512} & {\small 7.8135} & {\small 7.7283} & {\small 7.7352} \\
  & \textbf{$h_k/H_k$} & \textbf{0.8669} & \textbf{0.5058} & \textbf{0.1662} & \textbf{0.0220} & \textbf{0.0141} & \textbf{0.0150} \\
\midrule
\multirow{3}{*}{DeepSeek-R1 8.19B}
  & $h_k$      & {\small 5.4991} & {\small 4.3606} & {\small 2.6554} & {\small 1.3696} & {\small 1.2106} & {\small 1.3834} \\
  & $H_k$      & {\small 9.7100} & {\small 9.3782} & {\small 8.8974} & {\small 8.3018} & {\small 8.1619} & {\small 8.1946} \\
  & \textbf{$h_k/H_k$} & \textbf{0.5663} & \textbf{0.4650} & \textbf{0.2984} & \textbf{0.1650} & \textbf{0.1483} & \textbf{0.1688} \\
\midrule
\multirow{3}{*}{Qwen2.5 7.62B}
  & $h_k$      & {\small 6.3029} & {\small 4.7156} & {\small 2.4262} & {\small 0.7460} & {\small 0.4949} & {\small 0.5389} \\
  & $H_k$      & {\small 10.2246} & {\small 9.6059} & {\small 8.7706} & {\small 8.0662} & {\small 7.9394} & {\small 7.9347} \\
  & \textbf{$h_k/H_k$} & \textbf{0.6164} & \textbf{0.4909} & \textbf{0.2766} & \textbf{0.0925} & \textbf{0.0623} & \textbf{0.0679} \\
\bottomrule
\end{tabular}}
\end{table*}

\begin{table*}[ht]
\centering
\caption{Conditional entropy $h_k$, average‐distribution entropy $H_k$ (bits), and their ratio $h_k/H_k$  for each model on the \textit{Kant's Critique of Judgement} corpus.}
\label{tab:entropy_ratio_kant}
\resizebox{\textwidth}{!}{%
\begin{tabular}{l l r r r r r r}
\toprule
\textbf{Model} & \textbf{Metric} & \textbf{k=3} & \textbf{9} & \textbf{30} & \textbf{90} & \textbf{300} & \textbf{600} \\
\midrule
\multirow{3}{*}{Llama 3.3 70.6B}
  & $h_k$      & {\small 11.5800} & {\small 5.8695} & {\small 3.1620} & {\small 1.6728} & {\small 1.4296} & {\small 1.5319} \\
  & $H_k$      & {\small 13.4347} & {\small 10.2062} & {\small 8.7336} & {\small 8.0251} & {\small 7.8971} & {\small 7.8724} \\
  & \textbf{$h_k/H_k$} & \textbf{0.8619} & \textbf{0.5751} & \textbf{0.3621} & \textbf{0.2084} & \textbf{0.1810} & \textbf{0.1946} \\
\midrule
\multirow{3}{*}{DeepSeek-R1 8.19B}
  & $h_k$      & {\small 5.6830} & {\small 4.7242} & {\small 3.6155} & {\small 2.9959} & {\small 2.8928} & {\small 2.9508} \\
  & $H_k$      & {\small 9.8826} & {\small 9.4601} & {\small 8.8292} & {\small 8.4820} & {\small 8.3635} & {\small 8.3338} \\
  & \textbf{$h_k/H_k$} & \textbf{0.5751} & \textbf{0.4994} & \textbf{0.4095} & \textbf{0.3532} & \textbf{0.3459} & \textbf{0.3541} \\
\midrule
\multirow{3}{*}{Qwen2.5 7.62B}
  & $h_k$      & {\small 6.4208} & {\small 5.2071} & {\small 3.7593} & {\small 3.2637} & {\small 3.2129} & {\small 3.2450} \\
  & $H_k$      & {\small 10.3692} & {\small 9.6701} & {\small 8.7061} & {\small 8.4114} & {\small 8.3495} & {\small 8.3070} \\
  & \textbf{$h_k/H_k$} & \textbf{0.6192} & \textbf{0.5385} & \textbf{0.4318} & \textbf{0.3880} & \textbf{0.3848} & \textbf{0.3906} \\
\bottomrule
\end{tabular}}
\end{table*}

\paragraph{Overall Decline in Uncertainty}%
Across all model-corpus pairs, the uncertainty index \(u_k\) generally decreases as context length \(k\) grows, with the largest reductions occurring between short windows (e.g., \(k=3\) to \(k=30\)).  
At the longest windows (\(k=300\) to \(k=600\)), some runs exhibit small non-monotonic fluctuations, consistent with finite-sample variance in the Monte Carlo estimates.

\paragraph{Diminishing Returns from Longer Context}%
Although \(u_k\) falls steadily, its decline slows once the context becomes long.
The change from \(k=300\) to \(k=600\) is small and can be non-monotonic, consistent with finite-sample variance in the Monte Carlo estimates.
Thus most predictive benefit comes from early context, while very long windows add only modest gains.

\subsection{Model-Specific Cognitive Profiles}%
\begin{figure*}[t]
    \centering
    \includegraphics[width=0.8\textwidth]{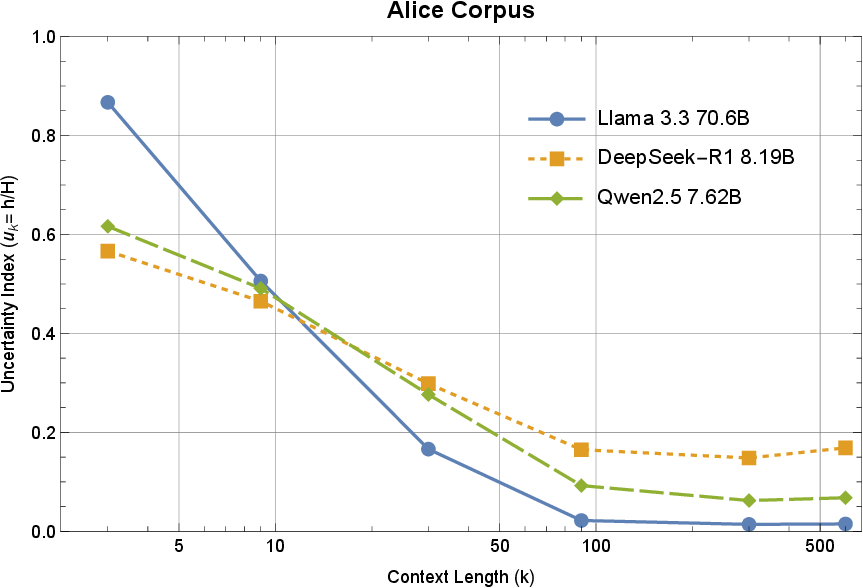}
    \caption{Entropy Decay Curves (\(u_k\) vs.\ \(k\)) on \textit{Alice's Adventures in Wonderland}.}
    \label{fig:Alice_decay_curves}
\end{figure*}

\begin{figure*}[t]
    \centering
    \includegraphics[width=0.8\textwidth]{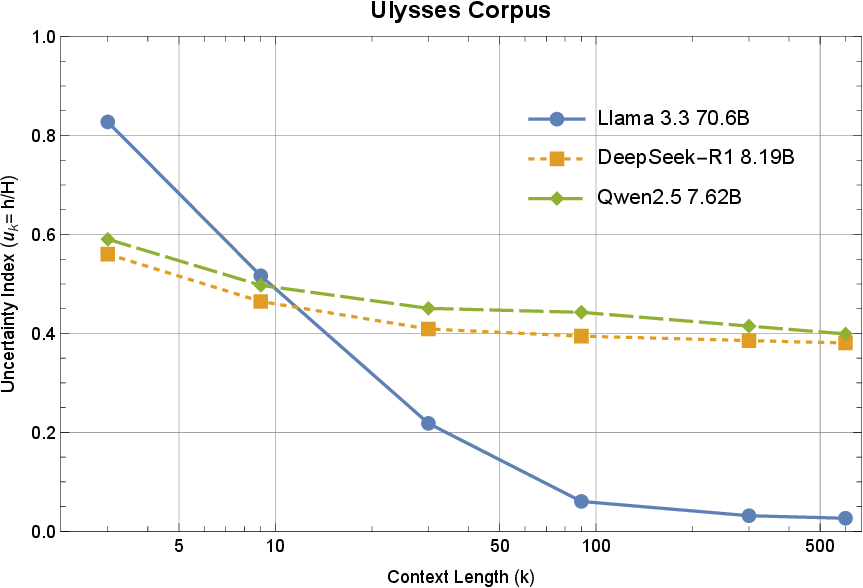}
    \caption{Entropy Decay Curves (\(u_k\) vs.\ \(k\)) on \textit{Ulysses}.}
    \label{fig:Ulysses_decay_curves}
\end{figure*}

\begin{figure*}[t]
    \centering
    \includegraphics[width=0.8\textwidth]{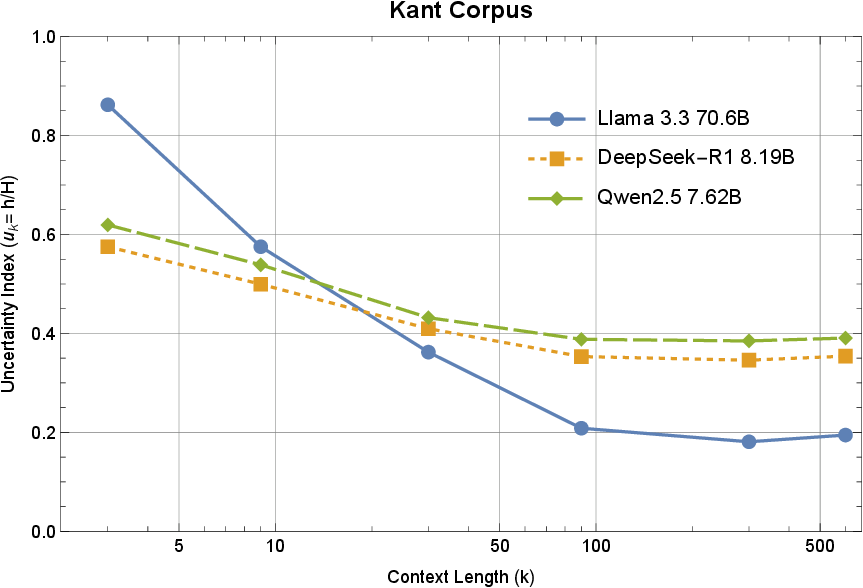}
    \caption{Entropy Decay Curves (\(u_k\) vs.\ \(k\)) on \textit{Kant's Critique of Judgement}.}
    \label{fig:Kant_decay_curves}
\end{figure*}

Figures~\ref{fig:Alice_decay_curves}--\ref{fig:Kant_decay_curves} plot the Entropy Decay Curves for each
model-corpus pair. The \(x\)-axis is logarithmic to emphasise behaviour at short context lengths.

\begin{itemize}
    \item \textbf{Large-scale model (Llama 3.3).}
          At very short contexts (\(k=3\)), Llama 3.3 exhibits the highest uncertainty across the tested corpora, indicating a broad set of plausible continuations. As \(k\) increases, its uncertainty decays much more rapidly than the smaller models, reaching the lowest long-context levels (notably on \textit{Alice} and \textit{Ulysses}). This sharp transition is consistent with stronger context utilisation at larger scale.

    \item \textbf{Smaller models (DeepSeek-R1 and Qwen 2.5).}
          Both smaller models show a gentler decline in \(u_k\). On challenging texts such as \textit{Ulysses} and \textit{Kant's Critique of Judgement}, their curves remain comparatively high even at long contexts, suggesting more limited long-range disambiguation.
\end{itemize}

\paragraph{Corpus-Dependent Profiles and Text Complexity}
The uncertainty index \(u_k\) depends strongly on the corpus. Across all three models,
\emph{Alice's Adventures in Wonderland} exhibits substantially lower uncertainty at moderate-to-long
context lengths (notably \(k\ge 30\)). This implies that once sufficient conditioning context is
available, the model's next-token distribution becomes sharply concentrated. By contrast,
\emph{Ulysses} and \emph{Kant's Critique of Judgement} maintain higher \(u_k\) over the same range,
indicating that considerable residual uncertainty persists even with long contexts. In this sense,
the Entropy Decay Curve (EDC) functions as an empirical \emph{predictability profile}: it quantifies
how effectively a model achieves contextual disambiguation across texts of differing effective
predictive difficulty.

\subsection{Detecting Memorisation and Data Contamination}
The EDC also flags potential memorisation.  
When Llama 3.3 is run on the \emph{Alice} corpus, its \(u_k\) values fall \emph{almost to zero} at long contexts, an extremity unlikely to arise from genuine generalisation.  
This is consistent with the hypothesis that large portions of \emph{Alice} appeared in its pre-training data.  
By contrast, on the \emph{Kant} corpus the same model keeps noticeably higher \(u_k\) values, consistent with real predictive processing.  
Thus the EDC can (i) audit test sets for contamination and (ii) separate models that truly generalise from those that rely mainly on memorised content.

\subsection{The Information Gain Span (IGS) as a Summary Metric}
To summarize the overall trajectory of an entropy decay curve in a single scalar, we introduce the \emph{Information Gain Span} (IGS):
\[
\text{IGS} \;:=\; u_{k_{\text{small}}} \cdot \bigl(1 - u_{k_{\text{large}}}\bigr),
\]
where $k_{\text{small}}$ and $k_{\text{large}}$ denote, respectively, short- and long-range context windows. Intuitively, a larger IGS reflects a desirable profile: high initial uncertainty (large $u_{k_{\text{small}}}$) that decays to strong certainty (small $u_{k_{\text{large}}}$). Table~\ref{tab:igs_results} reports IGS scores for all experiments using $k_{\text{small}} = 3$ and $k_{\text{large}} = 600$.

\begin{table}[!ht]
\centering
\caption{Information Gain Span (IGS) computed as $\mathrm{IGS}=u_{3}(1-u_{600})$. A higher IGS score indicates a better balance between high initial uncertainty and low final uncertainty.}
\label{tab:igs_results}
\begin{tabular}{@{}lccc@{}}
\toprule
\textbf{Model} & \textbf{Alice} & \textbf{Ulysses} & \textbf{Kant} \\
\midrule
Llama 3.3 70.6B       & 0.8539 & 0.8054 & 0.6942 \\
DeepSeek-R1 8.19B     & 0.4707 & 0.3470 & 0.3715 \\
Qwen2.5 7.62B         & 0.5745 & 0.3548 & 0.3773 \\
\bottomrule
\end{tabular}
\end{table}

The IGS scores quantitatively confirm our qualitative observations. The Llama 3.3 model achieves the highest IGS score on the \textit{Alice} and \textit{Ulysses} corpora, reflecting its superior ability to transition from divergent to convergent processing. IGS is highest on Alice for all models, while Ulysses and Kant yield lower and broadly comparable scores, with their ordering varying by model.

\section{Discussion}

The Entropy Decay Curves (EDCs) do more than rank models; they provide a direct, quantitative view of how large language models process information.  
Because each model-corpus pair yields a distinct curve, the results illuminate properties of both the models and the texts.  
We highlight two main implications.

\subsection{Predictability Profiles of Model-Text Interaction}

An EDC serves as a \textbf{predictability profile}: it shows how quickly a model's uncertainty falls as it reads a text.
For every model tested, \emph{Ulysses} and \emph{Kant's Critique of Judgement} yield higher uncertainty than \emph{Alice} for moderate-to-long contexts (especially \(k\ge 30\)), and therefore remain predictive challenges over longer spans.
The first two works therefore pose a longer-lasting predictive challenge.

Unlike traditional readability indices \cite{kincaid1975derivation}, this profile is data-driven.  
The model itself reveals a text's effective predictability through its evolving output distribution, without relying on hand-crafted complexity rules.

\subsection{Entropy Collapse as an Anomaly Signal}

The same method flags unusual behaviour.  
On the \emph{Alice} corpus, Llama 3.3 pushes the uncertainty index \(u_k\) almost to zero at long contexts---a phenomenon we call \textbf{entropy collapse}.  
Such extreme certainty is unlikely without exposure to the exact text and is therefore strong evidence of memorisation from training-data contamination \cite{carlini2022quantifying, achiam2023gpt}.  
Because \emph{Alice's Adventures in Wonderland} is a public-domain classic, its presence in pre-training data is plausible.

The pattern differs on the \emph{Kant} corpus, where Llama 3.3 maintains much higher uncertainty.  
Thus an entropy collapse on a benchmark text is a reliable warning sign: the model may be recalling rather than generalising, and the training data merit closer inspection.

\section{Theoretical Properties of the Entropy Metrics}
\label{sec:rigor}

This section provides a theoretical basis for the empirically observed patterns of \(h_k\) and \(H_k\).
We relate the finite-sample estimators of Section~\ref{sec:definitions} to corresponding population
quantities under a local sampling distribution over contexts.

\subsection{From Sample Estimates to Expected Values}
We now define population (theoretical) counterparts of the empirical quantities $h_k$ and $H_k$ by
replacing the finite-window average by an expectation under the same sliding-window sampling rule.

Fix a context length $k$ and consider the evaluated token sequence $(T_1,T_2,\dots)$.
Select a window-start index $I$ \emph{uniformly at random} from the admissible starts used by the
estimator (e.g.\ $I\in\{1,\dots,N\}$ in our experiments). Define the random context and target
\[
X^{(k)} := (T_I,\dots,T_{I+k-1}),
\qquad
Y^{(k)} := T_{I+k}.
\]
Given a realised context $X^{(k)}=x$, the model outputs next-token probabilities $p(y\mid x)$ for
$y\in\mathcal{Y}$. Using the same predictive-entropy definition as in
Section~\ref{sec:definitions}, define
\[
H(Y^{(k)}\mid X^{(k)}=x)
\;:=\;
-\sum_{y\in\mathcal{Y}} p(y\mid x)\,\log_2 p(y\mid x).
\]

\paragraph{Theoretical conditional entropy.}
We define the theoretical average conditional entropy at length $k$ as
\[
h_k \;:=\; \mathbb{E}\!\left[\,H\!\left(Y^{(k)}\mid X^{(k)}\right)\right]
\;=\;
\mathbb{E}\!\left[\,
-\sum_{y\in\mathcal{Y}} p\!\left(y\mid X^{(k)}\right)\log_2 p\!\left(y\mid X^{(k)}\right)
\right],
\]
where the expectation is over the random window start $I$ (equivalently, over the induced random
window $(X^{(k)},Y^{(k)})$).

\paragraph{Theoretical marginal entropy.}
Define the averaged predictive distribution
\[
\bar p_k(y) \;:=\; \mathbb{E}\!\left[p\!\left(y\mid X^{(k)}\right)\right],
\qquad y\in\mathcal{Y},
\]
and its entropy
\[
H_k \;:=\; -\sum_{y\in\mathcal{Y}} \bar p_k(y)\,\log_2 \bar p_k(y).
\]

\paragraph{Connection to the empirical estimators.}
The empirical quantities of Section~\ref{sec:definitions} are Monte Carlo estimates of these
expectations under the uniform sliding-window sampling rule: the sample mean $\frac{1}{N}\sum_i$
approximates $\mathbb{E}[\cdot]$, and the empirical average distribution
$\frac{1}{N}\sum_i p(y\mid x_i)$ approximates $\bar p_k(y)$.

\subsection{Monotonicity of the Average Conditional Entropy ($h_k$)}
\label{ssec:hk_monotonicity_revisited}
Empirically, $h_k$ decreases as $k$ grows. In our setting, $h_k$ and $h_{k+1}$ correspond to
different targets ($T_{I+k}$ versus $T_{I+k+1}$), so we use a local shift-invariance assumption to
compare them under the same window-averaging rule.

\begin{assumption}[Positional homogeneity / local shift-invariance]
\label{ass:positional_homogeneity}
Under the sliding-window sampling rule, the distribution of local blocks is (approximately)
invariant to a one-step shift: replacing every sampled block
$(T_I,\dots,T_{I+k})$ by $(T_{I-1},\dots,T_{I+k-1})$ does not materially change its distribution
within the evaluated segment.
\end{assumption}

\begin{proposition}[Monotonicity of $h_k$ under positional homogeneity]
\label{prop:hk_monotonicity_positional}
Assume Assumption~\ref{ass:positional_homogeneity}. With
\[
h_k := \mathbb{E}\!\left[\,H\!\left(T_{I+k}\mid T_I,\dots,T_{I+k-1}\right)\right],
\]
we have $h_{k+1}\le h_k$ (and, under approximate homogeneity, $h_{k+1}\lesssim h_k$).
\end{proposition}

\begin{proof}
By definition,
\[
h_{k+1}
=
\mathbb{E}\!\left[\,H\!\left(T_{I+k+1}\mid T_I,\dots,T_{I+k}\right)\right].
\]
By Assumption~\ref{ass:positional_homogeneity}, shifting indices by one gives
\[
h_{k+1}
\;\approx\;
\mathbb{E}\!\left[\,H\!\left(T_{I+k}\mid T_{I-1},\dots,T_{I+k-1}\right)\right].
\]
Conditioning reduces entropy pointwise, hence
\[
H\!\left(T_{I+k}\mid T_{I-1},\dots,T_{I+k-1}\right)
\;\le\;
H\!\left(T_{I+k}\mid T_I,\dots,T_{I+k-1}\right).
\]
Taking expectations yields $h_{k+1}\lesssim h_k$, and under exact shift-invariance the inequality is
$h_{k+1}\le h_k$.
\end{proof}

\paragraph{Connection to our estimator.}
For each realised window start $I=i$, the term
$H(T_{i+k}\mid T_i,\dots,T_{i+k-1})$ is exactly the predictive entropy computed from the model's
next-token distribution $p(y\mid T_i,\dots,T_{i+k-1})$ as in Section~\ref{sec:definitions}. Averaging
over $I$ matches the sliding-window averaging used to compute the empirical $h_k$.

\subsection{Monotonicity of the Marginal Entropy (\(H_k\))}
Models trained on natural language often exhibit a decline in \(H_k\) as \(k\) grows, but this is
not an information-theoretic necessity: \(H_k\) is an entropy of an \emph{average} predictive
distribution, and mixture effects can in principle make it increase or stay flat.

\begin{remark}[Why \(H_k\) Depends on \(k\)]
\label{remark:Hk_dependence}
Recall our theoretical definition from the previous subsection:
\[
\bar p_k(y) := \mathbb{E}\!\left[p\!\left(y\mid X^{(k)}\right)\right],
\qquad
H_k := -\sum_{y\in\mathcal{Y}} \bar p_k(y)\,\log_2 \bar p_k(y).
\]
Thus \(H_k\) can depend on \(k\) through two mechanisms.

\begin{enumerate}
\item \textbf{Context-length effect.} Increasing \(k\) changes the conditioning information in
\(p(y\mid x)\), which can systematically reshape the averaged distribution \(\bar p_k\).

\item \textbf{Sampling-position effect.} In a finite evaluated segment, the sliding-window rule
defines \(X^{(k)}=(T_I,\dots,T_{I+k-1})\) with targets at positions \(I+k\). Changing \(k\) shifts the
set of target positions being averaged over, so the induced distribution of contexts \(X^{(k)}\) (and
hence \(\bar p_k\)) may change even if the underlying text is non-stationary.
\end{enumerate}

Our blockwise positional robustness checks (Section~\ref{ssec:positional_robustness}) indicate that,
within the tested \emph{Alice} segment, the estimates of \(h_k\), \(H_k\), and \(u_k\) remain stable
across widely separated blocks (for both \(k=3\) and \(k=600\) in our updated table), suggesting that
the sampling-position effect is small in that setting.
\end{remark}

\paragraph{Decomposition of Predictive Uncertainty.}
Under the sliding-window sampling scheme, a randomly sampled context \(X^{(k)}\) induces
next-token probabilities \(p(y\mid X^{(k)})\) for \(y\in\mathcal{Y}\).
Averaging these predictive distributions over the sampling rule yields the
\emph{averaged predictive distribution}
\[
\bar p_k(y) := \mathbb{E}\!\left[p\!\left(y\mid X^{(k)}\right)\right], \qquad y\in\mathcal{Y},
\]
whose entropy is
\[
H_k := -\sum_{y\in\mathcal{Y}} \bar p_k(y)\,\log_2 \bar p_k(y).
\]
Conversely, the quantity
\[
h_k := \mathbb{E}\!\left[
-\sum_{y\in\mathcal{Y}} p\!\left(y\mid X^{(k)}\right)\log_2 p\!\left(y\mid X^{(k)}\right)
\right]
\]
is the expected predictive entropy \emph{within} a randomly sampled context.

We define the information-gain term
\[
I_k := H_k - h_k,
\]
which measures the expected reduction in uncertainty when predicting with the
context-specific distribution \(p(\cdot\mid X^{(k)})\) rather than with the averaged
distribution \(\bar p_k\).
By Jensen's inequality (concavity of entropy), \(H_k \ge h_k\), hence \(I_k\ge 0\),
and the decomposition \(H_k = h_k + I_k\) holds by definition.

\subsection{Relationship Between the Decay Rates of \(h_k\) and \(H_k\)}
Define single-step differences \(\Delta h_k := h_k - h_{k+1}\) and
\(\Delta H_k := H_k - H_{k+1}\).
Under Proposition~\ref{prop:hk_monotonicity_positional} we typically have \(\Delta h_k\ge 0\),
whereas \(\Delta H_k\) need not be nonnegative in general (though it is often observed to be so in
practice).

\begin{remark}[Two Regimes of Uncertainty Reduction]
When \(\Delta H_k>0\), comparing \(\Delta h_k\) and \(\Delta H_k\) is a useful diagnostic:
\begin{enumerate}
    \item \textbf{Local-prediction dominance (\(\Delta h_k > \Delta H_k\)).}  
          The token greatly helps predict the very next symbol, so \(h_k\) drops sharply, but it does little to reshape the long-term distribution, so \(H_k\) falls less.
    \item \textbf{Global-structure dominance (\(\Delta H_k > \Delta h_k\)).}  
          The token clarifies the topic or style, strongly narrowing the whole future space.  
          The \textbf{Global Structure Dominance Condition} is \(\Delta H_k \ge \Delta h_k\).
\end{enumerate}
Which regime applies can vary with model, corpus, and \(k\).  
Comparing these decay rates is therefore a useful diagnostic.
\end{remark}

\subsection{A Symmetric View via Mutual Information}
\begin{proposition}[Symmetric decomposition of \(H_k-h_k\)]
Fix \(k\) and let \(X^{(k)}\) be the random context under the sliding-window sampling rule as in the
previous subsection. Define an auxiliary next-token random variable \(\tilde Y^{(k)}\) by
\[
\Pr(\tilde Y^{(k)}=y \mid X^{(k)}=x) := p(y\mid x), \qquad y\in\mathcal{Y}.
\]
Then the marginal law of \(\tilde Y^{(k)}\) is \(\Pr(\tilde Y^{(k)}=y)=\bar p_k(y)\), hence
\[
H_k = H(\tilde Y^{(k)}), \qquad h_k = H(\tilde Y^{(k)}\mid X^{(k)}).
\]
Consequently,
\[
H_k - h_k
= I\!\left(X^{(k)};\tilde Y^{(k)}\right)
= H\!\left(X^{(k)}\right) - H\!\left(X^{(k)}\mid \tilde Y^{(k)}\right).
\]
\end{proposition}

\begin{proof}
By construction, \(H_k = H(\tilde Y^{(k)})\) and \(h_k = H(\tilde Y^{(k)}\mid X^{(k)})\), so
\(H_k-h_k = I(X^{(k)};\tilde Y^{(k)})\).
Mutual information is symmetric:
\[
I(A;B)=H(A)-H(A\mid B)=H(B)-H(B\mid A).
\]
Applying this with \(A=X^{(k)}\) and \(B=\tilde Y^{(k)}\) yields the stated identity.
\end{proof}

\paragraph{Interpretation.}
\(H_k-h_k\) is the expected information gain from conditioning on a specific context rather than
using the averaged predictive distribution. The symmetric form shows it is equally the expected
information gained about the sampled context from observing a token drawn from the model-induced
next-token distribution.

\subsection{Theoretical Analysis of the Information-Gain Span}
\label{ssec:igs_theory_revised}
Let a short and long context window be \(k_{\mathrm{s}}\ll k_{\mathrm{l}}\).  
Recall the \emph{Information-Gain Span}
\[
  \operatorname{IGS}(k_{\mathrm{s}},k_{\mathrm{l}})
  := u_{k_{\mathrm{s}}}\bigl(1-u_{k_{\mathrm{l}}}\bigr),
  \qquad u_k := h_k/H_k.
\]

\begin{theorem}[IGS and Markov order (regime separation)]
\label{thm:optimal_revised}
Assume the text source is an ergodic Markov chain of exact order \(m\) and the model predicts perfectly.  
Assume further that \(H_k\) is non-increasing.  
Then \(u_k\) is non-increasing for \(k<m\) and non-decreasing for \(k\ge m\) (it may be flat for \(k\ge m\)).  
In particular, \(u_k\) attains a minimum at \(k=m\), though the minimiser need not be unique.
\end{theorem}

\begin{proof}
\textbf{Step 1: Behaviour of \(u_k\).}  
For \(k<m\), \(h_k\) falls strictly and \(H_k\) does not rise, so \(u_k\) falls.  
For \(k\ge m\), \(h_k\) is constant (\(h_m\)) while \(H_k\) is non-increasing, so \(u_k=h_m/H_k\) is non-decreasing or stays flat.  
Thus \(u_k\) attains a minimum at \(k=m\) (the minimiser need not be unique if \(H_k\) is flat for \(k\ge m\)).

\textbf{Step 2: The factor \((1-u_{k_{\mathrm{l}}})\).}
This term is largest when \(u_{k_{\mathrm{l}}}\) is smallest.
Under the theorem's conclusion, \(u_k\) attains its minimum at \(k=m\) (and may remain flat on an
interval). Hence the maximal value of \((1-u_{k_{\mathrm{l}}})\) is achieved by choosing
\(k_{\mathrm{l}}\) at a minimiser of \(u_k\), i.e.\ \(k_{\mathrm{l}}=m\) (or any \(k_{\mathrm{l}}\ge m\)
if \(u_k\) is flat for \(k\ge m\)).
In practice, one may still choose a larger \(k_{\mathrm{l}}\) to represent a ``long-context'' regime,
even when this does not improve the idealised bound.

\textbf{Step 3: The factor \(u_{k_{\mathrm{s}}}\).}  
Because \(u_k\) falls on \(k<m\), \(u_{k_{\mathrm{s}}}\) is largest when \(k_{\mathrm{s}}\) is as small as allowed, subject to \(k_{\mathrm{s}}\le m\).
\end{proof}

\paragraph{Interpretation.}
Under a finite-memory source, \(u_k\) decreases until the relevant history is captured (around \(m\)) and then ceases to improve.  
This motivates selecting \(k_{\mathrm{s}} \le m < k_{\mathrm{l}}\) so that IGS contrasts a pre-saturation short-range regime with a post-saturation long-range regime.  
Using IGS peaks to estimate an effective memory scale is therefore heuristic and would require additional assumptions for a formal guarantee.

\section{Conclusion}

We proposed a task-agnostic framework for probing how large language models process information by
tracking predictive uncertainty as context grows. The core object is the \emph{Entropy Decay Curve}
(EDC), defined by plotting the length-conditional uncertainty index \(u_k=h_k/H_k\) across context
lengths \(k\). Unlike single-number summaries, the EDC exposes a model's full trajectory from
short-context ambiguity to long-context stabilisation, yielding a quantitative \emph{cognitive
profile}.

Across three open models and three public-domain corpora, we observed stable and interpretable
differences in these profiles. Larger models tended to show a sharper transition from high initial
uncertainty to low long-context uncertainty, while smaller models often exhibited a slower decay and
a higher plateau on more difficult texts. The same analysis revealed strong corpus effects: texts
that become locally predictable with sufficient context produced markedly lower long-context
uncertainty than stylistically or semantically demanding works.

To summarise a decay pattern with a single scalar, we introduced the \emph{Information Gain Span}
(IGS), which contrasts a short-context regime with a long-context regime. Empirically, IGS tracked
the qualitative desirability of a profile---high initial uncertainty coupled with strong long-range
resolution---and enabled direct comparison across model--corpus pairs.

Beyond comparison, EDCs also act as a diagnostic. In particular, near-zero long-context uncertainty
(\emph{entropy collapse}) is difficult to reconcile with genuine generalisation on natural text and
therefore serves as a practical warning signal for memorisation or benchmark contamination.

Finally, we connected the estimators to population quantities under the same sliding-window sampling
rule and provided sufficient conditions under which \(h_k\) is monotone in \(k\). We also clarified
that monotonicity of \(H_k\) is not guaranteed in general, and interpreted \(H_k-h_k\) as an expected
information gain under the induced context distribution.

\paragraph{Limitations and future work.}
Our study is limited to a small set of models, corpora, and window lengths, and relies on
finite-sample Monte Carlo estimates. Future work should (i) extend coverage to more architectures and
training regimes, (ii) evaluate specialised domains (code, scientific, medical, legal), (iii)
characterise estimator variance and sensitivity to sampling design, and (iv) refine contamination
tests by combining EDC signals with independent memorisation audits.

\section*{Acknowledgements}

This study relies exclusively on publicly released large language models.  
We thank the developers and the wider open-source community for providing these resources.

All experiments used three models: a 70.6B-parameter member of the Llama 3 family, DeepSeek-R1, and Qwen 2.5.  
In accordance with the licence of the first model, we state: \emph{Built with Llama}.

\noindent\textbf{Licensing.}
\begin{itemize}
    \item \textbf{Llama 3.3}: distributed under the \emph{Llama 3.3 Community License}.  
          Copyright~\copyright~Meta Platforms, Inc.\ All rights reserved.
    \item \textbf{DeepSeek-R1}: released under the MIT Licence.
    \item \textbf{Qwen 2.5}: released under the Apache Licence, Version 2.0.
\end{itemize}

Our use of each model complies fully with its licence and acceptable-use policy.

This work was partially supported by the KIST Institutional Program.

\bibliography{references}

@article{neisser1979concept,
  title={The concept of intelligence},
  author={Neisser, Ulric},
  journal={Intelligence},
  volume={3},
  number={3},
  pages={217--227},
  year={1979},
  publisher={Elsevier}
}

@article{sternberg2004culture,
  title={Culture and intelligence.},
  author={Sternberg, Robert J},
  journal={American psychologist},
  volume={59},
  number={5},
  pages={325},
  year={2004},
  publisher={American Psychological Association}
}

@article{shannon1948,
	author = {Shannon, C. E.},
	journal = {The Bell System Technical Journal},
	number = {3},
	pages = {379-423},
	title = {A mathematical theory of communication},
	volume = {27},
	year = {1948}}

@article{qwen3,
    title={Qwen3 Technical Report}, 
    author={An Yang and Anfeng Li and Baosong Yang and Beichen Zhang and Binyuan Hui and Bo Zheng and Bowen Yu and Chang Gao and Chengen Huang and Chenxu Lv and Chujie Zheng and Dayiheng Liu and Fan Zhou and Fei Huang and Feng Hu and Hao Ge and Haoran Wei and Huan Lin and Jialong Tang and Jian Yang and Jianhong Tu and Jianwei Zhang and Jianxin Yang and Jiaxi Yang and Jing Zhou and Jingren Zhou and Junyang Lin and Kai Dang and Keqin Bao and Kexin Yang and Le Yu and Lianghao Deng and Mei Li and Mingfeng Xue and Mingze Li and Pei Zhang and Peng Wang and Qin Zhu and Rui Men and Ruize Gao and Shixuan Liu and Shuang Luo and Tianhao Li and Tianyi Tang and Wenbiao Yin and Xingzhang Ren and Xinyu Wang and Xinyu Zhang and Xuancheng Ren and Yang Fan and Yang Su and Yichang Zhang and Yinger Zhang and Yu Wan and Yuqiong Liu and Zekun Wang and Zeyu Cui and Zhenru Zhang and Zhipeng Zhou and Zihan Qiu},
    journal = {arXiv preprint arXiv:2505.09388},
    year={2025}
}

@article{qwen2.5,
    title   = {Qwen2.5 Technical Report}, 
    author  = {An Yang and Baosong Yang and Beichen Zhang and Binyuan Hui and Bo Zheng and Bowen Yu and Chengyuan Li and Dayiheng Liu and Fei Huang and Haoran Wei and Huan Lin and Jian Yang and Jianhong Tu and Jianwei Zhang and Jianxin Yang and Jiaxi Yang and Jingren Zhou and Junyang Lin and Kai Dang and Keming Lu and Keqin Bao and Kexin Yang and Le Yu and Mei Li and Mingfeng Xue and Pei Zhang and Qin Zhu and Rui Men and Runji Lin and Tianhao Li and Tingyu Xia and Xingzhang Ren and Xuancheng Ren and Yang Fan and Yang Su and Yichang Zhang and Yu Wan and Yuqiong Liu and Zeyu Cui and Zhenru Zhang and Zihan Qiu},
    journal = {arXiv preprint arXiv:2412.15115},
    year    = {2024}
}

@article{bommasani2021opportunities,
  title={On the opportunities and risks of foundation models},
  author={Bommasani, Rishi and Hudson, Drew A and Adeli, Ehsan and Altman, Russ and Arora, Simran and von Arx, Sydney and Bernstein, Michael S and Bohg, Jeannette and Bosselut, Antoine and Brunskill, Emma and others},
  journal={arXiv preprint arXiv:2108.07258},
  year={2021}
}

@inproceedings{bender2021dangers,
  title={On the dangers of stochastic parrots: Can language models be too big?},
  author={Bender, Emily M and Gebru, Timnit and McMillan-Major, Angelina and Shmitchell, Shmargaret},
  booktitle={Proceedings of the 2021 ACM conference on fairness, accountability, and transparency},
  pages={610--623},
  year={2021}
}

@article{kadavath2022language,
  title={Language models (mostly) know what they know},
  author={Kadavath, Saurav and Conerly, Tom and Askell, Amanda and Henighan, Tom and Drain, Dawn and Perez, Ethan and Schiefer, Nicholas and Hatfield-Dodds, Zac and DasSarma, Nova and Tran-Johnson, Eli and others},
  journal={arXiv preprint arXiv:2207.05221},
  year={2022}
}

@article{jiang2020can,
  title={How can we know what language models know?},
  author={Jiang, Zhengbao and Xu, Frank F and Araki, Jun and Neubig, Graham},
  journal={Transactions of the Association for Computational Linguistics},
  volume={8},
  pages={423--438},
  year={2020},
  publisher={MIT Press One Rogers Street, Cambridge, MA 02142-1209, USA journals-info~…}
}

@article{tishby2000information,
  title={The information bottleneck method},
  author={Tishby, Naftali and Pereira, Fernando C and Bialek, William},
  journal={arXiv preprint physics/0004057},
  year={2000}
}

@article{shwartz2017opening,
  title={Opening the black box of deep neural networks via information},
  author={Shwartz-Ziv, Ravid and Tishby, Naftali},
  journal={arXiv preprint arXiv:1703.00810},
  year={2017}
}

@book{goodfellow2016deep,
  title={Deep learning},
  author={Goodfellow, Ian and Bengio, Yoshua and Courville, Aaron and Bengio, Yoshua},
  volume={1},
  number={2},
  year={2016},
  publisher={MIT Press},
  address={Cambridge}
}

@article{vaswani2017attention,
  title={Attention is all you need},
  author={Vaswani, Ashish and Shazeer, Noam and Parmar, Niki and Uszkoreit, Jakob and Jones, Llion and Gomez, Aidan N and Kaiser, {\L}ukasz and Polosukhin, Illia},
  journal={Advances in neural information processing systems},
  volume={30},
  year={2017}
}

@book{cover1999elements,
  title={Elements of information theory},
  author={Cover, Thomas M},
  year={1999},
  publisher={John Wiley \& Sons},
  address={New York}
}

@techreport{kincaid1975derivation,
  title={Derivation of new readability formulas (automated readability index, fog count and flesch reading ease formula) for navy enlisted personnel},
  author={Kincaid, J Peter and Fishburne Jr, Robert P and Rogers, Richard L and Chissom, Brad S},
  year={1975}
}

@inproceedings{carlini2022quantifying,
  title={Quantifying memorization across neural language models},
  author={Carlini, Nicholas and Ippolito, Daphne and Jagielski, Matthew and Lee, Katherine and Tramer, Florian and Zhang, Chiyuan},
  booktitle={The Eleventh International Conference on Learning Representations},
  year={2022}
}

@article{achiam2023gpt,
  title={Gpt-4 technical report},
  author={Achiam, Josh and Adler, Steven and Agarwal, Sandhini and Ahmad, Lama and Akkaya, Ilge and Aleman, Florencia Leoni and Almeida, Diogo and Altenschmidt, Janko and Altman, Sam and Anadkat, Shyamal and others},
  journal={arXiv preprint arXiv:2303.08774},
  year={2023}
}

@article{brown2020language,
  title={Language models are few-shot learners},
  author={Brown, Tom and Mann, Benjamin and Ryder, Nick and Subbiah, Melanie and Kaplan, Jared D and Dhariwal, Prafulla and Neelakantan, Arvind and Shyam, Pranav and Sastry, Girish and Askell, Amanda and others},
  journal={Advances in neural information processing systems},
  volume={33},
  pages={1877--1901},
  year={2020}
}

@book{bishop2006pattern,
  title={Pattern recognition and machine learning},
  author={Bishop, Christopher M and Nasrabadi, Nasser M},
  volume={4},
  number={4},
  year={2006},
  publisher={Springer},
  address={New York}
}

\end{document}